\documentclass{article}
\usepackage{ijcai24}

\usepackage{times}
\usepackage{soul}
\usepackage{url}
\usepackage[hidelinks]{hyperref}
\usepackage[utf8]{inputenc}
\usepackage[small]{caption}
\usepackage{graphicx}
\usepackage{amsmath}
\usepackage{amsthm}
\usepackage{booktabs}
\usepackage{algorithm}
\usepackage{algorithmic}
\usepackage[switch]{lineno}

\usepackage[permil]{overpic}
\usepackage{xcolor}         % colors
\usepackage{arydshln}  
\usepackage{amsmath}
\usepackage{amssymb}
\usepackage{booktabs} 
\usepackage{multirow}

\newcommand\ie{i.e.,~}
\newcommand\et{\textit{et al.~}}
\newcommand\tb{\textbf}
\newcommand{\ours}[0]{\textsc{EliMipl}}
\newcommand{\ourssp}[0]{\textsc{Sp}}
\newcommand{\oursin}[0]{\textsc{In}}
\newcommand{\oursma}[0]{\textsc{Ma}}

\newcommand{\demipl}[0]{\textsc{DeMipl}}
\newcommand{\miplgp}[0]{\textsc{MiplGp}}
\newcommand{\proden}[0]{\textsc{Proden}}
\newcommand{\rc}[0]{\textsc{Rc}}
\newcommand{\lws}[0]{\textsc{Lws}}
\newcommand{\aggd}[0]{\textsc{Pl-aggd}}

\definecolor{steelblue}{RGB}{70, 130, 180}
\title{Exploiting Conjugate Label Information for Multi-Instance \\Partial-Label Learning}

\author{
Wei Tang$^{1,2}$
\and
Weijia Zhang$^3$\and
Min-Ling Zhang$^{1,2}$\thanks{Corresponding author}\\
\affiliations
$^1$School of Computer Science and Engineering, Southeast University, Nanjing {\rm 210096}, China\\
$^2$Key Lab. of Computer Network and Information Integration (Southeast University), MoE, China\\
$^3$School of Information and Physical Sciences, The University of Newcastle, NSW {\rm 2308}, Australia \\
\emails
tangw@seu.edu.cn, \space 
weijia.zhang@newcastle.edu.au, \space 
zhangml@seu.edu.cn
}

\begin{document}

\maketitle

\begin{abstract}
	Multi-instance partial-label learning (MIPL) addresses scenarios where each training sample is represented as a multi-instance bag associated with a candidate label set containing one true label and several false positives. Existing MIPL algorithms have primarily focused on mapping multi-instance bags to candidate label sets for disambiguation, disregarding the intrinsic properties of the label space and the supervised information provided by non-candidate label sets. In this paper, we propose an algorithm named {\ours}, \ie \textit{Exploiting conjugate Label Information for Multi-Instance Partial-Label learning}, which exploits the conjugate label information to improve the disambiguation performance. To achieve this, we extract the label information embedded in both candidate and non-candidate label sets, incorporating the intrinsic properties of the label space. Experimental results obtained from benchmark and real-world datasets demonstrate the superiority of the proposed {\ours} over existing MIPL algorithms and other well-established partial-label learning algorithms. 
\end{abstract}

\section{Introduction}
\label{sec:int}
Weakly supervised learning has emerged as a powerful strategy in scenarios with limited annotated data. Based on label quality and quantity, weak supervision can be broadly categorized into three types: inaccurate, inexact, and incomplete supervision \cite{zhou2018brief}. Inexact supervision refers to a coarse correspondence between instances and labels. To work with inexact supervision, these are two prevalent learning paradigms, \ie \emph{multi-instance learning} (MIL) \cite{amores2013multiple,carbonneau2018multiple,IlseTW18,Weijia22,zhang2022dtfd} and \emph{partial-label learning} (PLL) \cite{cour2011learning,lyu2020hera,ZhangF0L0QS22,HeFLLY22,GongYB22,LiJLWO23}.
In MIL, a sample is represented as a \emph{bag of instances} and associated with a \emph{single bag-level label}, while the instance-level labels are inaccessible to the learner. In PLL, a sample is represented as a \emph{single instance} and linked to a \emph{candidate label set}, including one true label and multiple false positives. Therefore, MIL and PLL can be perceived as two sides of the same coin: inexact supervision within MIL manifests in the instance space, whereas inexact supervision appears in the label space within PLL. 
\begin{figure}[!t]
\centering
\begin{overpic}[width=85mm]{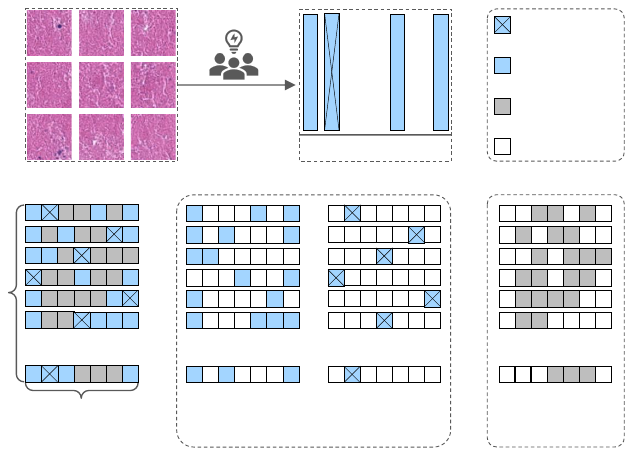}   
\put(480, 470) {\scriptsize $k_1$}
\put(513, 470) {\scriptsize $k_2$}
\put(548, 470) {\scriptsize $k_3$}
\put(583, 470) {\scriptsize $k_4$}
\put(617, 470) {\scriptsize $k_5$}
\put(652, 470) {\scriptsize $k_6$}
\put(687, 470) {\scriptsize $k_7$}
\put(298, 540) {\scriptsize crowd-sourced}
\put(292, 510) {\scriptsize candidate labels}
\put(822, 666) {\scriptsize ground-truth}
\put(822, 635) {\scriptsize labels}
\put(822, 599) {\scriptsize false positive}
\put(822, 569) {\scriptsize labels}
\put(822, 537) {\scriptsize non-candidate}
\put(822, 505) {\scriptsize labels}
\put(822, 472) {\scriptsize zero entries} 
\put(365, 425) {\small \tb{(a)}}
\put(838, 425) {\small legends} 
\put(-8, 240) {\small $m$} 
\put(240, 240) {\small $=$} 
\put(487, 240) {\small $+$} 
\put(732, 240) {\small $+$} 
\put(130, 143) {\small \rotatebox{90}{$\cdots$}} 
\put(382, 143) {\small \rotatebox{90}{$\cdots$}} 
\put(608, 143) {\small \rotatebox{90}{$\cdots$}} 
\put(878, 143) {\small \rotatebox{90}{$\cdots$}} 
\put(91, 60) {\small $k=7$} 
\put(67, 20) {\small complete} 
\put(30, -20) {\small label matrix {\scriptsize{$\boldsymbol{Y}$}}}
\put(322, 20) {\small candidate label matrix {\scriptsize{$\boldsymbol{S}$}}} 
\put(779, 60) {\small non-candidate} 
\put(779, 20) {\small label matrix {\scriptsize{$\boldsymbol{\bar{S}}$}}} 
\put(370, 65) {\small {\scriptsize{$\boldsymbol{Y_F}$}}} 
\put(526, 65) {\small {\scriptsize{$\boldsymbol{Y_T}$}} (\emph{sparse})} 
\put(486, -32) {\small \tb{(b)}} 
\end{overpic}
\caption{\tb{(a)} A multi-instance bag is labeled with a candidate label set $\mathcal{S}=\{k_1, k_2, k_5, k_7\}$. \tb{(b)} The decomposition of the complete label matrix, where $m$ and $k$ represent the number of multi-instance bags and categories, respectively.} 
  \label{fig:ill}
\end{figure}

However, many tasks exhibit a phenomenon of \emph{dual inexact supervision}, where ambiguity arises in both instance and label spaces. To work with the dual inexact supervision, Tang \et \shortcite{tang2023miplgp} introduced a learning paradigm known as \emph{multi-instance partial-label learning} (MIPL) and developed a Gaussian Processes-based algorithm ({\miplgp}), which derives a bag-level predictor by aggregating predictions of all instances within the same bag. To capture global representations for multi-instance bags, an algorithm named {\demipl} equipped with an attention mechanism is introduced \cite{tang2023demipl}. The existing algorithms mainly operate in the instance space and only utilize the candidate label information.

The non-candidate label set holds crucial roles in MIPL. In histopathological image classification, images are commonly segmented into patches \cite{campanella2019}, and their labels may come from crowd-sourced annotators rather than expert pathologists \cite{GroteSFWF19}. Figure \ref{fig:ill}(a) illustrates that crowd-sourced annotators treat an image as a multi-instance bag $\boldsymbol{X}_i = \{\boldsymbol{x}_{i,1}, \boldsymbol{x}_{i,2}, \cdots, \boldsymbol{x}_{i,9}\}$ and provide a candidate label set $\mathcal{S}_i = \{k_1, k_2, k_5, k_7\}$, whose candidate label matrix can be written as $\boldsymbol{S}_i = [1,1,0,0,1,0,1]$. Similarly, the non-candidate label set $\mathcal{\bar{S}}_i = \{k_3, k_4, k_6\}$ corresponds to the non-candidate label matrix $\boldsymbol{\bar{S}}_i = [0,0,1,1,0,1,0]$, indicating that $\boldsymbol{X}_i$ must not belong to categories $k_3$, $k_4$, or $k_6$. Therefore, we can extract exact supervision from the non-candidate label set. As depicted in Figure \ref{fig:ill}(b), we decompose a complete label matrix $\boldsymbol{Y}$ into a candidate label matrix $\boldsymbol{S}$ and a non-candidate label matrix $\boldsymbol{\bar{S}}$. Subsequently, $\boldsymbol{S}$ may be further disintegrated into a false positive label matrix $\boldsymbol{Y_F}$ and a true label matrix $\boldsymbol{Y_T}$, \ie $\boldsymbol{Y} = \boldsymbol{S} + \boldsymbol{\bar{S}} = \boldsymbol{Y_F} + \boldsymbol{Y_T} + \boldsymbol{\bar{S}}$. Notably, $\boldsymbol{Y_T}$ is sparse, as each row must have one and only one non-zero element. However, the current MIPL algorithms have predominantly concentrated on the mappings from multi-instance bags to $\boldsymbol{S}$, neglecting the sparsity of $\boldsymbol{Y_T}$ and the information from $\boldsymbol{\bar{S}}$.
\begin{figure}[!h]
    \centering
    \includegraphics[width=80mm]{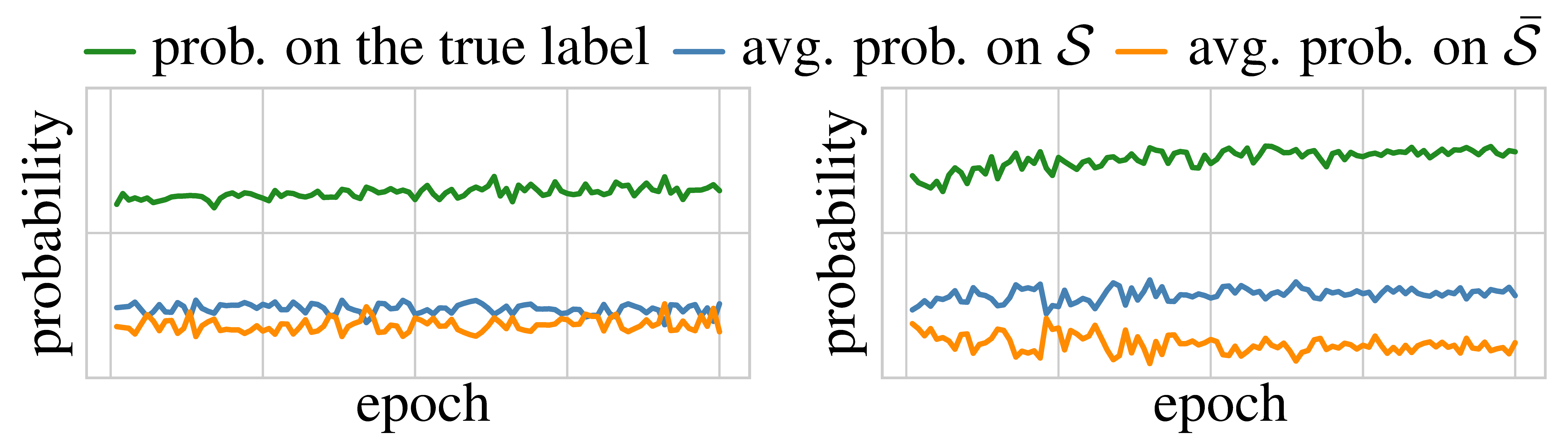}
   \caption{Predicted probabilities of {\demipl} (left) and {\ours} (right) on the sample in CRC-{\scriptsize{MIPL-Row}} dataset. }   
\label{fig:prob}
\end{figure}

Consequently, Figure \ref{fig:prob} illustrates the predicted probabilities on the true label, along with the average predicted probabilities on each candidate label and non-candidate label. The left side depicts the probabilities of the {\demipl}, revealing proximity in the average predicted probabilities on candidate and non-candidate labels. This observation indicates that \textit{{\demipl} encounters difficulty in effectively discerning between candidate and non-candidate labels}. To address this challenge, we introduce the concept of \emph{conjugate label information} (CLI), encapsulating information from both candidate and non-candidate label sets, along with the sparsity of the true label matrix. The right side in Figure \ref{fig:prob} shows the predicted probabilities when exploiting the CLI. It is evident that (a) the predicted probabilities on the true label exhibit a noticeable increase, (b) the average predicted probabilities on the non-candidate label are reduced, and (c) the average probabilities on each candidate label and non-candidate label are distinctly separated. This suggests that \textit{the CLI conduce to train a more discriminative MIPL classifier}.

In this paper, we present an algorithm named {\ours}, i.e., \textit{Exploiting conjugate Label Information for Multi-Instance Partial-Label learning}. Firstly, we introduce a scaled additive attention mechanism to aggregate each multi-instance bag into a bag-level feature representation. Secondly, to enhance the utilization of candidate label information, we leverage the mappings from the bag-level features to the candidate label sets, coupled with the sparsity of the candidate label matrix. Lastly, to incorporate the non-candidate label information, we propose an inhibition loss to diminish the model's predictions on the non-candidate labels. To the best of our knowledge, we are the first to introduce the scaled additive attention mechanism and the CLI in MIPL. Extensive experimental results demonstrate that {\ours} outperforms the state-of-the-art MIPL algorithms and the PLL algorithms.

The remainder is organized as follows. Firstly, we review related work in Section \ref{sec:related}. Secondly, we present the proposed {\ours} in Section \ref{sec:methodology} and report the experimental results in Section \ref{sec:experiments}. Lastly, we conclude this paper in Section \ref{sec:con}.
\begin{figure*}[!t]
\centering
\begin{overpic}[width=140mm]{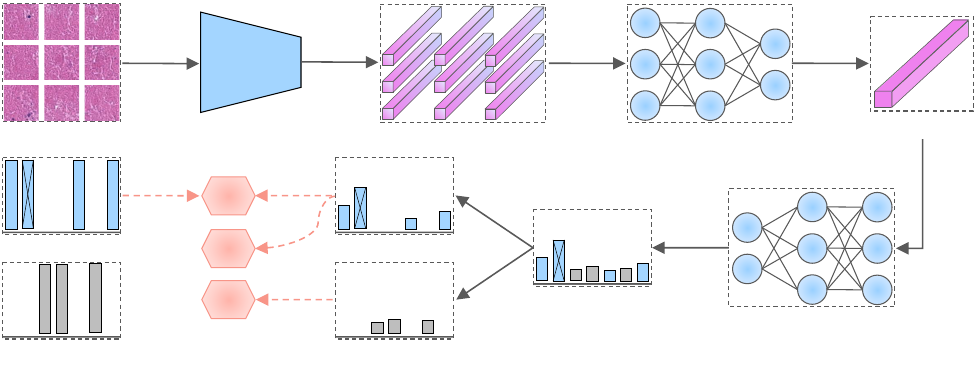}   
\put(-35, 232) {\small multi-instance bag $\boldsymbol{X}_i$}
\put(235, 310) {\small $\psi(\cdot)$}
\put(180, 232) {\small feature extractor}
\put(360, 232) {\small instance-level features $\boldsymbol{H}_i$}
\put(620, 232) {\small scaled additive attention}
\put(855, 245) {\small bag-level feature $\boldsymbol{z}_i$}
\put(-30, 120) {\small candidate label set $\mathcal{S}_i$}
\put(-50, 12) {\small non-candidate label set $\mathcal{\bar{S}}_i$}
\put(218, 66) {\small $\mathcal{L}_{\text{in}}$}
\put(218, 120) {\small $\mathcal{L}_{\text{sp}}$}
\put(218, 173) {\small $\mathcal{L}_{\text{ma}}$}
\put(324, 120) {\small probabilities on $\mathcal{S}_i$}
\put(324, 12) {\small probabilities on $\mathcal{\bar{S}}_i$}
\put(533, 66) {\small probabilities $\boldsymbol{P}_i$}
\put(795, 45) {\small classifier}
\end{overpic}
\caption{The pipeline of {\ours}, where $\mathcal{L}_{\text{ma}}$, $\mathcal{L}_{\text{sp}}$, and $\mathcal{L}_{\text{in}}$ refer to mapping loss,  sparsity loss, and inhibition loss, respectively.}
  \label{fig:framework}
\end{figure*}
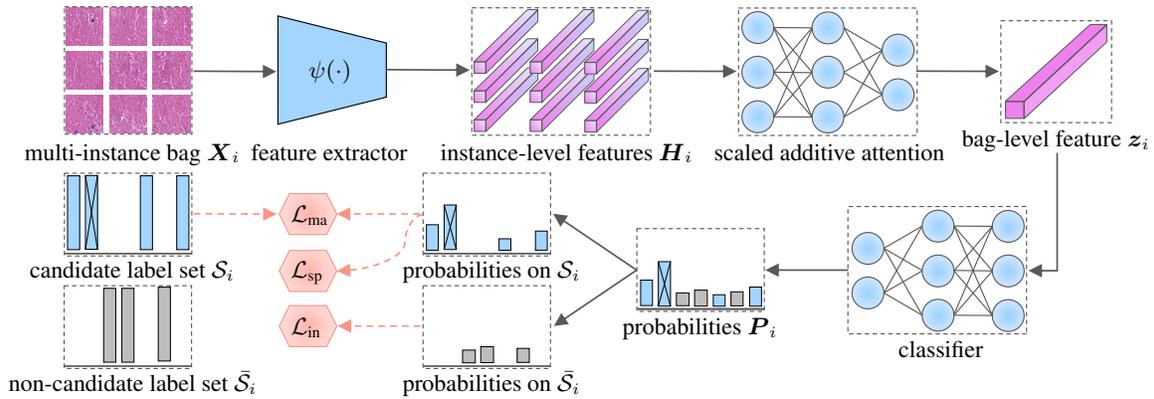

\section{Related Work}
\label{sec:related}
\subsection{Multi-Instance Learning}
Originating from drug activity prediction \cite{dietterich1997}, MIL has found extensive adoption in diverse applications, including text classification \cite{ZhouSL09,Weijia21} and image annotation \cite{wang2018revisiting}. Contemporary deep MIL approaches predominantly rely on attention mechanisms \cite{WangPWW22,ChenYO22,TanZSW23}. Ilse \et \shortcite{IlseTW18} introduced attention mechanisms to aggregate each multi-instance bag into a feature vector. For multi-classification tasks, Shi \et \shortcite{Shi20} proposed a loss-based attention mechanism to learn instance-level weights, predictions, and bag-level predictions. Furthermore, researchers have explored the intrinsic attributes of attention mechanisms to improve performance \cite{cui2023bayesmil,xiang2023exploring}. While these approaches achieve promising results in cases with exact bag-level labels, they face challenges in learning from ambiguous bag-level labels.

\subsection{Partial-Label Learning}
Recent PLL approaches heavily rely on deep learning techniques. Yao \et \shortcite{YaoDC0W020} employed deep convolutional neural networks for feature extraction and utilized the exponential moving average technique to uncover latent true labels. Building on the empirical risk minimization principle, Lv \et \shortcite{LvXF0GS20} devised a classifier-consistent risk estimator that progressively identifies true labels. Similarly, Feng \et \shortcite{FengL0X0G0S20} delved into the generation process of partial-labeled data, proposing both a risk-consistent approach and a classifier-consistent approach. Taking a more generalized stance, Wen \et \shortcite{WenCHL0L21} presented a weighted loss function capable of accommodating various methods through distinct weight assignments. Furthermore, Wu \et \shortcite{WuWZ22} proposed a supervised loss to constrain outputs on non-candidate labels, coupled with consistency regularization on candidate labels. While the supervised loss bears resemblance to our inhibition loss, our proposed CLI loss incorporates additional components, namely the mapping loss and the sparse loss. Although these methods effectively learn from partial-labeled data, they lack the capability to manage multi-instance bags.

\subsection{Multi-Instance Partial-Label Learning}
In contrast to the inherent limitations of addressing only unilateral inexact supervision in MIL and PLL, MIPL possesses the capability to work with dual inexact supervision. To the best of our knowledge, there are only two viable MIPL algorithms. Tang \et \shortcite{tang2023miplgp} is the first to introduce the framework of MIPL along with a Gaussian processes-based algorithm ({\miplgp}), which follows an instance-space paradigm. {\miplgp} begins by augmenting a negative class for each candidate label set, subsequently treating the candidate label set of each multi-instance bag as that of each instance within the bag. Finally, it employs the Dirichlet disambiguation strategy and the Gaussian processes regression model for disambiguation. Differing from {\miplgp}, {\demipl} follows the embedded-space paradigm and aggregates each multi-instance bag into a feature representation and employs a momentum-based disambiguation strategy to find true labels from candidate label sets \cite{tang2023demipl}. However, both methods primarily depend on mapping from instances or multi-instance bags to candidate label sets for disambiguation, without considering the proposed CLI in this paper.

\section{Methodology}
\label{sec:methodology}
\subsection{Preliminaries}
In this study, we define a MIPL training dataset as $\mathcal{D} = \{(\boldsymbol{X}_i, \mathcal{S}_i) \mid 1 \le i \le m\}$, comprising $m$ multi-instance bags and their corresponding candidate label sets. Specifically, a candidate label set $\mathcal{S}_i$ consists of one true label and multiple false positive labels, but the true label is unknown. It is crucial to note that a bag contains at least one instance pertaining to the true label, while excluding any instances corresponding to false positive labels. The instance space is denoted as $\mathcal{X} \in \mathbb{R}^d$, while the label space $\mathcal{Y} = \{1, 2, \cdots, k\}$ encompasses $k$ class labels. The $i$-th bag $\boldsymbol{X}_i = \{\boldsymbol{x}_{i,1}, \boldsymbol{x}_{i,2}, \cdots, \boldsymbol{x}_{i,n_i}\}$ comprises $n_i$ instances of dimension $d$. Both the candidate label set $\mathcal{S}_i$ and the non-candidate label set $\bar{\mathcal{S}}_i$ are proper subsets of the label space $\mathcal{Y}$, satisfying the conditions $|\mathcal{S}_i| + |\bar{\mathcal{S}}_i| = |\mathcal{Y}| = k$, where $|\cdot|$ denotes the cardinality of a set. 

The pipeline of the proposed {\ours} is depicted in Figure \ref{fig:framework}, which contains three main components: an instance-level feature extractor, a scaled additive attention mechanism, and a classifier. When presented with a multi-instance bag $\boldsymbol{X}_i$ along with its associated candidate label set $\mathcal{S}_i$ and non-candidate label set $\bar{\mathcal{S}}_i$, we initially employ a feature extractor to procure instance-level feature representations. Subsequently, the scaled additive attention mechanism is applied to aggregate a bag of instances into a unified bag-level feature representation. Finally, the classifier is invoked to estimate the class probabilities based on the bag-level features. To utilize the CLI, we introduce a mapping loss $\mathcal{L}_{\text{ma}}$ and a sparsity loss $\mathcal{L}_{\text{sp}}$ to disambiguate the candidate label sets, along with an inhibition loss $\mathcal{L}_{\text{in}}$ to suppress the model’s prediction over the non-candidate label sets.

\subsection{Instance-Level Feature Extractor}
For a given multi-instance bag $\boldsymbol{X}_i = \{\boldsymbol{x}_{i,1}, \boldsymbol{x}_{i,2}, \cdots, \boldsymbol{x}_{i,n_i}\}$ with $n_i$ instances, instance-level feature representations $\boldsymbol{H}_i$ are learned using a feature extractor $\psi(\cdot)$ as follows:
\begin{equation}
\label{eq:extractor}
	\boldsymbol{H}_i = \psi(\boldsymbol{X}_i) = \{\boldsymbol{h}_{i, 1},  \boldsymbol{h}_{i, 2},  \cdots,  \boldsymbol{h}_{i, n_i}\},
\end{equation}
where $\boldsymbol{h}_{i,j} \in \mathbb{R}^{l}$ indicates the feature representation of the $j$-th instance within the $i$-th multi-instance bag, and $\psi(\cdot)$ is a neural network comprised of two components, \ie $\psi(\boldsymbol{X}_i) = \psi_2(\psi_1(\boldsymbol{X}_i))$. Here, $\psi_1(\cdot)$ is a feature extractor that can be tailored to the specific characteristics of the datasets, and $\psi_2(\cdot)$ is composed of fully connected layers that map instance-level features to an embedded space of dimension $l$.

\subsection{Scaled Additive Attention Mechanism}
To aggregate instance-level features into bag-level representations, we introduce a scaled additive attention mechanism specifically designed for MIPL. The existing attention mechanism for MIPL utilizes the sigmoid function for calculating attention scores, followed by normalization \cite{tang2023demipl}. The attention scores derived through the sigmoid function are constrained within the range $(0, 1)$, leading to a limited distinction between instances. Therefore, we introduce an additive attention mechanism calculating attention scores by the softmax function to distinguish instances, equipped with a scaling factor to prevent vanishing gradients \cite{VaswaniSPUJGKP17}. Specifically, we first denote the output of the additive attention mechanism as $\xi(h_{i, j})$, quantifying the impact of the $j$-th instance on the $i$-th bag as follows:
\begin{equation}
\label{eq:contribution}
    	\xi (h_{i, j}) = \boldsymbol{W}^\top (\text{tanh}(\boldsymbol{W}_{t}^\top \boldsymbol{h}_{i, j} + \boldsymbol{b}_t) \odot \text{sigm}(\boldsymbol{W}_{s}^\top \boldsymbol{h}_{i, j} + \boldsymbol{b}_s)),
\end{equation}
where $\boldsymbol{W}^\top$, $\boldsymbol{W}_{t}^\top$, $\boldsymbol{W}_{s}^\top$, $\boldsymbol{b}_{t}$, and $\boldsymbol{b}_{s}$ are learnable parameters. $\text{tanh}(\cdot)$ and $\text{sigm}(\cdot)$ are the hyperbolic tangent and sigmoid functions, respectively. The operator $\odot$ denotes element-wise multiplication. 
Then, we normalize $\xi(h_{i, j})$ using softmax with a scaling factor $1/\sqrt{l}$ to derive the attention score: 
\begin{equation}
\label{eq:attention}
	a_{i, j} = \frac{\exp \left(\xi \left(h_{i, j}\right) / \sqrt{l} \right)} {\sum_{j{^\prime}=1}^{n_i} \exp \left(\xi \left(h_{i, j^\prime}\right) / \sqrt{l} \right)},
\end{equation}
where $a_{i,j}$ represents the attention score of the $j$-th instance in the $i$-th bag. Finally, we consolidate the instance-level features into a bag-level representation, as demonstrated below: 
\begin{equation}
\label{eq:aggregation}
	\boldsymbol{z}_i = \sum_{j=1}^{n_i} a_{i, j} \boldsymbol{h}_{i, j},
\end{equation}
where $\boldsymbol{z}_i$ represents the bag-level representation of the $i$-th multi-instance bag. The bag-level representations of all multi-instance bags in the training dataset are denoted by $\mathcal{Z}$.

\subsection{Conjugate Label Information}
\paragraph{Candidate Label Information.} Once the bag-level feature representations are acquired, the subsequent task is to disambiguate the candidate label set. The disambiguation entails establishing the mapping relationship from the bag-level features to their corresponding candidate label set. The goal of precise mapping is to guide the classifier to assign higher class probabilities to true labels and lower probabilities to false positive labels. To attain this objective, we employ a weighted mapping loss function:
\begin{equation}
\label{eq:mapping}
	\mathcal{L}_{\text{ma}}(\mathcal{Z},  \mathcal{S}) = -\frac{1}{m} \sum_{i=1}^{m} \sum_{c \in \mathcal{S}_i} w_{i, c}^{(t)} \log (f_c(\boldsymbol{z}_i)),
\end{equation}
where $f$ is the classifier, and $f_{c}(\cdot)$ represents the classifier's prediction probability for the candidate label $c$. $w_{i, c}^{(t)}$ denotes the weight assigned to the prediction of the $c$-th class at the $t$-th epoch, using the features of the $i$-th bag as input for the classifier. For candidate labels, we initialize $w_{i,c}^{(0)}=\frac{1}{\left|\mathcal{S}_i\right|}$ through an averaging approach. During training, we update $w_{i,c}^{(t)}$ by computing a weighted sum of the classifier's outputs at both the previous epoch and current epoch as follows:
\begin{equation}
\label{eq:update_w}
	w_{i,c}^{(t)} = \rho^{(t)} w_{i,c}^{(t-1)}  + (1-\rho^{(t)}) \frac{f_c(\boldsymbol{z}_i)}{\sum_{c^\prime \in \mathcal{S}_i} f_{c^\prime}(\boldsymbol{z}_i)},
\end{equation} 
where $\rho^{(t)} = {(T-t)}/{T}$ is dynamically adjusted across epochs, and $T$ is the maximum of the training epochs. 

While the mapping loss can assess the relative labeling probabilities of candidate labels, it fails to capture the mutually exclusive relationships among the candidate labels. To address this issue in PLL, Feng  \et \shortcite{feng2019partial} introduced the maximum infinity norm on the predicted probabilities of all classes and alternately optimize the maximum infinity norm by solving $k$ independent quadratic programming problems. However, as depicted in Figure \ref{fig:ill}(b), we observe that each row of the true label matrix exhibits sparsity. Although the true labels remain inaccessible during the training process, we encourage the classifier to generate sparse prediction probabilities for the candidate labels. Specifically, the goal is to push the prediction probability of the unknown true label toward $1$ while simultaneously driving the prediction probabilities of other candidate labels toward $0$. Therefore, we directly capture the mutually exclusive relationships among the candidate labels by implementing the sparsity loss, as detailed below:
\begin{equation}
\label{eq:sparsity}
	\mathcal{L}_{\text{sp}}(\mathcal{S}) = \frac{1}{m} \sum_{i=1}^{m} \|\boldsymbol{P}_i \odot \boldsymbol{S}_i\|_0,
\end{equation}
where $\boldsymbol{P}_i$ and $\boldsymbol{S}_i$ is the prediction probabilities and the candidate label set matrix of the $i$-th bag, respectively. $\odot$ denotes element-wise multiplication. Since minimizing the $\ell_0$ norm is NP-hard, we employ the $\ell_1$ norm as a surrogate for the $\ell_0$ norm, promoting sparsity while allowing for efficient optimization \cite{wright2022high}.

\begin{algorithm}[!t] 
\caption{Training Procedure of {\ours}}
\label{alg:algorithm}
\textbf{Inputs}: \\
$\mathcal{D}$ : MIPL training set $\{(\boldsymbol{X}_i,\mathcal{S}_i) \mid 1 \le i \le m \}$\\ 
$\mu$, $\gamma$ : Weights for sparsity loss and inhibition loss \\
$T$: Maximum number of epochs \\
\textbf{Process}: 
\begin{algorithmic}[1]  
\STATE Initialize uniform weights $w_{i,c}^{(0)}$ ($c \in \mathcal{S}_i$)
\FOR{$t=1$ to $T$}
	\STATE Fetch a mini-batch $\mathcal{B}$ from $\mathcal{D}$
   	 \FOR{$\boldsymbol{X} \in \mathcal{B}$}
        \STATE Extract instance-level features using Equation (\ref{eq:extractor}) \\
        \STATE Calculate attention scores using Equations (\ref{eq:contribution}, \ref{eq:attention})  \\
        \STATE Aggregate instance-level features into bag-level feature representations via Equation (\ref{eq:aggregation}) \\
        \STATE Update weights $w_{i,c}^{(t)}$ based on Equation (\ref{eq:update_w}) \\
        \STATE Calculate $\mathcal{L}_{\text{ma}}$, $\mathcal{L}_{\text{sp}}$, and $\mathcal{L}_{\text{in}}$ via Equations (\ref{eq:mapping}, \ref{eq:sparsity}, \ref{eq:inhibition}) \\
        \STATE Calculate total loss $\mathcal{L}$ as in Equation (\ref{eq:full_loss}) \\
        \STATE Set gradient $- \bigtriangledown_{\Phi} \mathcal{L}$ \\
        \STATE Update $\Phi$ using optimizer \\
\ENDFOR
\ENDFOR
\end{algorithmic}
\end{algorithm}

\begin{table*}[!t] 
\centering \scriptsize 
\begin{tabular}{p{4.3cm} | p{2.5 cm} p{2.5 cm}  p{1.9 cm} p{1.9 cm} p{1.9 cm}p{1.9 cm}p{1.9 cm} p{2.5 cm} p{2.9 cm}}
\hline \hline
	\multicolumn{1}{l|}{Dataset}				& \multicolumn{1}{c}{\#bag}		& \multicolumn{1}{c}{\#ins} 		& \multicolumn{1}{c}{max. \#ins}		& \multicolumn{1}{c}{min. \#ins}			& \multicolumn{1}{c}{avg. \#ins}	 	& \multicolumn{1}{c}{\#dim}		& \multicolumn{1}{c}{\#class} 		& \multicolumn{1}{c}{avg. \#CLs }	& \multicolumn{1}{c}{domain}		\\ 	
	\hline
MNIST-{\scriptsize{MIPL}} (MNIST) 				& \multicolumn{1}{c}{500}			& \multicolumn{1}{c}{20664}		& \multicolumn{1}{c}{48}				& \multicolumn{1}{c}{35}				& \multicolumn{1}{c}{41.33}		& \multicolumn{1}{c}{784}			& \multicolumn{1}{c}{5}			& \multicolumn{1}{c}{{2, 3, 4}}		& \multicolumn{1}{c}{image}		\\ 
FMNIST-{\scriptsize{MIPL}} (FMNIST)			& \multicolumn{1}{c}{500}			& \multicolumn{1}{c}{20810}		& \multicolumn{1}{c}{48}				& \multicolumn{1}{c}{36}				& \multicolumn{1}{c}{41.62} 		&  \multicolumn{1}{c}{784}			& \multicolumn{1}{c}{5}			& \multicolumn{1}{c}{{2, 3, 4}}		& \multicolumn{1}{c}{image}		\\ 
Birdsong-{\scriptsize{MIPL}} (Birdsong)			& \multicolumn{1}{c}{1300}		& \multicolumn{1}{c}{48425}		& \multicolumn{1}{c}{76}				& \multicolumn{1}{c}{25}				& \multicolumn{1}{c}{37.25}		& \multicolumn{1}{c}{38}			& \multicolumn{1}{c}{13}			& \multicolumn{1}{c}{{2, 3, 4}}		& \multicolumn{1}{l}{biology}	\\ 
SIVAL-{\scriptsize{MIPL}} (SIVAL) 				& \multicolumn{1}{c}{1500}		& \multicolumn{1}{c}{47414}		& \multicolumn{1}{c}{32}				& \multicolumn{1}{c}{31}				& \multicolumn{1}{c}{31.61} 		& \multicolumn{1}{c}{30}			& \multicolumn{1}{c}{25}			& \multicolumn{1}{c}{{2, 3, 4}}		& \multicolumn{1}{c}{image}		\\ 
	  \hline 
CRC-{\scriptsize{MIPL-Row}} (C-Row)			& \multicolumn{1}{c}{7000}		& \multicolumn{1}{c}{56000}		& \multicolumn{1}{c}{8}				& \multicolumn{1}{c}{8}				& \multicolumn{1}{c}{8}			& \multicolumn{1}{c}{9}			& \multicolumn{1}{c}{7}			& \multicolumn{1}{c}{2.08}			& \multicolumn{1}{c}{image}	\\
CRC-{\scriptsize{MIPL-SBN}} (C-SBN)			& \multicolumn{1}{c}{7000}		& \multicolumn{1}{c}{63000}		& \multicolumn{1}{c}{9}				& \multicolumn{1}{c}{9}				& \multicolumn{1}{c}{9}			& \multicolumn{1}{c}{15}			& \multicolumn{1}{c}{7}			& \multicolumn{1}{c}{2.08}			& \multicolumn{1}{c}{image}\\
CRC-{\scriptsize{MIPL-KMeansSeg}} (C-KMeans) 	& \multicolumn{1}{c}{7000}		& \multicolumn{1}{c}{30178}		& \multicolumn{1}{c}{6}				& \multicolumn{1}{c}{3}				& \multicolumn{1}{c}{4.311}		& \multicolumn{1}{c}{6}			& \multicolumn{1}{c}{7}			& \multicolumn{1}{c}{2.08}			& \multicolumn{1}{c}{image}\\
CRC-{\scriptsize{MIPL-SIFT}} (C-SIFT)			& \multicolumn{1}{c}{7000}		& \multicolumn{1}{c}{175000}		& \multicolumn{1}{c}{25}				& \multicolumn{1}{c}{25}				& \multicolumn{1}{c}{25}			&\multicolumn{1}{c}{128}			& \multicolumn{1}{c}{7}			& \multicolumn{1}{c}{2.08}			& \multicolumn{1}{c}{image}	\\
	  \hline \hline
  \end{tabular}
    \caption{Characteristics of the benchmark and real-world MIPL datasets.}
    \label{tab:datasets}
\end{table*}

\paragraph{Non-candidate Label Information.} For a multi-instance bag $\boldsymbol{X}_i$ linked to a candidate label set $\mathcal{S}_i$, the non-candidate label set $\bar{\mathcal{S}}_i$ complements the candidate label set $\mathcal{S}_i$ within the label space $\mathcal{Y}$. As the label space has a fixed size, an antagonistic relationship arises between the non-candidate and candidate label sets. To enhance the classifier's prediction probabilities for the candidate label set, a natural strategy is to diminish the classifier's prediction probabilities for the non-candidate label set. Motivated by this insight, we introduce an inhibition loss as follows:
\begin{equation}
\label{eq:inhibition}
	\mathcal{L}_{\text{in}}(\mathcal{Z},  \bar{\mathcal{S}}) = -\frac{1}{m} \sum_{i=1}^{m} \sum_{\bar{c} \in \bar{\mathcal{S}_i}} \log (1 - f_{\bar{c}}(\boldsymbol{z}_i)),
\end{equation}
where $f_{\bar{c}}(\cdot)$ denotes the classifier's prediction probability over the non-candidate label $\bar{c}$. 

\paragraph{CLI Loss.} During the training, CLI is formed by a loss function named CLI loss that is a weighted fusion of the mapping loss, sparsity loss, and inhibition loss, as shown below:
\begin{equation}
\label{eq:full_loss}
	\mathcal{L} = \mathcal{L}_{\text{ma}}(\mathcal{Z},  \mathcal{S})  + \mu \mathcal{L}_{\text{sp}}(\mathcal{S}) + \gamma \mathcal{L}_{\text{in}}(\mathcal{Z},  \bar{\mathcal{S}}), 
\end{equation}
where $\mu$ and $\gamma$ represent the weighting coefficients for the sparsity loss and the inhibition loss, respectively.

Algorithm \ref{alg:algorithm} summarizes the training procedure of {\ours}. Firstly, the algorithm initializes the weights for the mapping loss uniformly (Step $1$). Subsequently, instance-level features are extracted and aggregated into bag-level features within each mini-batch (Steps $5$-$8$). The algorithm then updates the weights for the mapping loss and calculates the total loss function (Steps $9$-$11$). Finally, the model is optimized using gradient descent (Steps $12$ and $13$).

\section{Experiments}
\label{sec:experiments}
In this section, we begin by introducing the experimental configurations, including the datasets, comparative algorithms, and the parameters used in the experiments. Subsequently, we present the experimental results on both benchmark and real-world datasets. Finally, we conduct further analysis to gain deeper insights into the impact of CLI.

\subsection{Experimental Configurations}
\label{subsec:configurations}
\paragraph{Datasets.}
We employ four benchmark MIPL datasets \cite{tang2023miplgp,tang2023demipl}: MNIST-{\scriptsize{MIPL}}, FMNIST-{\scriptsize{MIPL}}, Birdsong-{\scriptsize{MIPL}}, and SIVAL-{\scriptsize{MIPL}}, spanning diverse domains such as image analysis and biology  \cite{lecun1998gradient,han1708,briggs2012rank,SettlesCR07}. The characteristics of the datasets are presented in Table \ref{tab:datasets}, where the abbreviations within parentheses in the first column represent the abbreviated names of the MIPL datasets. The dataset includes quantities of multi-instance bags and total instances, denoted as \emph{\#bag} and \emph{\#ins}, respectively. Additionally, we use \emph{max. \#ins}, \emph{min. \#ins}, and \emph{avg. \#ins} to indicate the maximum, minimum, and average instance count within all bags. The dimensionality of the instance-level feature is represented by \emph{\#dim}. Labeling details are elucidated using \emph{\#class} and \emph{avg. \#CLs}, signifying the length of the label space and the average length of candidate label sets, respectively. For a comprehensive performance assessment, we vary the count of false positive labels, denoted as $r$ ($|\mathcal{S}_i| = r + 1$).

CRC-{\scriptsize{MIPL}} dataset is a real-world MIPL dataset for colorectal cancer classification. We utilize multi-instance features generated by four image bag generators \cite{WeiZ16}: Row \cite{MaronR98}, single blob with neighbors (SBN) \cite{MaronR98}, k-means segmentation (KMeansSeg) \cite{ZhangGYF02}, and scale-invariant feature transform (SIFT) \cite{Lowe04}. 

The appendix contains detailed information about the datasets and the four image bag generators.

\paragraph{Comparative Algorithms.}
We conduct a comprehensive comparison involving {\ours} along with two established MIPL algorithms: {\miplgp} \cite{tang2023miplgp} and {\demipl} \cite{tang2023demipl}. These represent the entirety of available MIPL methods. Furthermore, we include four PLL algorithms: {\proden} \cite{LvXF0GS20}, {\rc} \cite{FengL0X0G0S20}, {\lws} \cite{WenCHL0L21}, and {\aggd} \cite{wangdb2022}. The first three algorithms can be equipped with diverse backbone networks, such as linear models and MLP. Due to spatial constraints, we present the results obtained from the linear models in the main body, while the results with MLP are shown in the appendix. Parameters for all algorithms are selected based on recommendations from original literature or refined through our search for enhanced outcomes.

Since PLL algorithms are not directly tailored for MIPL data, two common strategies, known as the Mean strategy and the MaxMin strategy, are employed to adapt MIPL data for PLL algorithms \cite{tang2023miplgp}. The Mean strategy involves calculating average feature values across all instances within a bag, resulting in a bag-level feature representation. In contrast, the MaxMin strategy identifies both the maximum and minimum feature values for each dimension among instances within a bag, and then concatenates these values to form a bag-level feature representation.

\paragraph{Implementation.}
We implement {\ours} using PyTorch and execute it on a single NVIDIA Tesla V100 GPU. We utilize the stochastic gradient descent (SGD) optimizer with a momentum value of $0.9$ and a weight decay of $0.0001$. The initial learning rate is selected from the set $\{0.01, 0.05\}$ and accompanied by a cosine annealing technique. We set the number of epochs uniformly to $100$ for all datasets. For the MNIST-{\scriptsize{MIPL}} and FMNIST-{\scriptsize{MIPL}} datasets, $\mu$ is set to $1$ or $0.1$, $\gamma$ is chosen from $\{0.1, 0.5\}$, and the feature extraction network $\psi_1(\cdot)$ is a two-layer convolutional neural network. For the remaining datasets, we set both $\mu$ and $\gamma$ to $10$, and $\psi_1(\cdot)$ is an identity transformation. The feature transformation network $\psi_2(\cdot)$ is implemented by a fully connected network, with the dimension $l$ set to $512$ for the CRC-{\scriptsize{MIPL}} dataset and $128$ for the other datasets. The way of dataset partitioning is consistent with that of {\demipl}. We conduct ten random train/test splits with a ratio of $7:3$. We report the mean accuracies and standard deviations obtained from the ten runs, with the highest accuracy highlighted in bold. The code of {\ours} can be found at \url{https://github.com/tangw-seu/ELIMIPL}.

\begin{table}[!t]
\centering 
  \scriptsize 
     \begin{tabular}{p{1.3cm} |c|cccc p{0.8cm}  p{136cm}  p{1.1cm}  p{1.1cm} p{1.1cm} }
    \hline  \hline
    \multicolumn{1}{l|}{Algorithm} 	& $r$ 	& MNIST 					& FMNIST 				& Birdsong 				& SIVAL \\
    \hline
    \multicolumn{1}{l|}{\multirow{3}[1]{*}{{\ours}}}
    							& 1     	& \textbf{.992$\pm$.007} 		& \textbf{.903$\pm$.018} 		& \textbf{.771$\pm$.018}		& \textbf{.675$\pm$.022} \\
          						& 2  		& \textbf{.987$\pm$.010} 		& \textbf{.845$\pm$.026} 		& \textbf{.745$\pm$.015} 		& \textbf{.616$\pm$.025} \\
          						& 3     	& \textbf{.748$\pm$.144} 		& \textbf{.702$\pm$.055} 		& \textbf{.717$\pm$.017} 		& \textbf{.600$\pm$.029} \\
     \hline
    \multicolumn{1}{l|}{\multirow{3}[1]{*}{{\demipl}}}
    							& 1     	& .976$\pm$.008	& .881$\pm$.021 	& .744$\pm$.016	& .635$\pm$.041 \\
          						& 2    	& .943$\pm$.027	& .823$\pm$.028	& .701$\pm$.024	& .554$\pm$.051\\
          						& 3     	& .709$\pm$.088	& .657$\pm$.025	& .696$\pm$.024	 & .503$\pm$.018\\
     \hline
   	\multicolumn{1}{l|}{\multirow{3}[1]{*}{{\miplgp}}} 
							& 1		& .949$\pm$.016	& .847$\pm$.030	& .716$\pm$.026	& .669$\pm$.019 \\
							& 2		& .817$\pm$.030	& .791$\pm$.027	& .672$\pm$.015	& .613$\pm$.026 \\
							& 3		& .621$\pm$.064	& .670$\pm$.052 	& .625$\pm$.015	& .569$\pm$.032\\
     \hline
    \multicolumn{6}{c}{Mean} \\
     \hline
    \multicolumn{1}{l|}{\multirow{3}[1]{*}{{\proden}}}  
    		  					& 1     	& .605$\pm$.023 	& .697$\pm$.042 	& .296$\pm$.014 	& .219$\pm$.014 \\
          						& 2     	& .481$\pm$.036 	& .573$\pm$.026 	& .272$\pm$.019	& .184$\pm$.014 \\
          						& 3     	& .283$\pm$.028 	& .345$\pm$.027 	& .211$\pm$.013 	& .166$\pm$.017 \\
     \hline
    \multicolumn{1}{l|}{\multirow{3}[1]{*}{{\rc}}} 
    		  					& 1     	& .658$\pm$.031 	& .753$\pm$.042 	& .362$\pm$.015 	& .279$\pm$.011 \\
          						& 2     	& .598$\pm$.033 	& .649$\pm$.028	& .335$\pm$.011 	& .258$\pm$.017 \\
          						& 3     	& .392$\pm$.033 	& .401$\pm$.063 	& .298$\pm$.009 	& .237$\pm$.020 \\
     \hline
   	\multicolumn{1}{l|}{\multirow{3}[1]{*}{{\lws}}} 
    		  					& 1     	& .463$\pm$.048 	& .726$\pm$.031 	& .265$\pm$.010 	& .240$\pm$.014 \\
          						& 2     	& .209$\pm$.028 	& .720$\pm$.025 	& .254$\pm$.010 	& .223$\pm$.008 \\
          						& 3     	& .205$\pm$.013 	& .579$\pm$.041 	& .237$\pm$.005 	& .194$\pm$.026 \\
     \hline
    \multicolumn{1}{l|}{\multirow{3}[1]{*}{{\aggd}}} 
    		  					& 1     	& .671$\pm$.027 	& .743$\pm$.026 	& .353$\pm$.019 	& .355$\pm$.015 \\
          						& 2     	& .595$\pm$.036 	& .677$\pm$.028 	& .314$\pm$.018 	& .315$\pm$.019 \\
          						& 3     	& .380$\pm$.032 	& .474$\pm$.057 	& .296$\pm$.015 	& .286$\pm$.018 \\
     \hline
    \multicolumn{6}{c}{MaxMin} \\
     \hline
    \multicolumn{1}{l|}{\multirow{3}[1]{*}{{\proden}}}  
    		 		 			& 1     	& .508$\pm$.024 	& .424$\pm$.045 	& .387$\pm$.014 	& .316$\pm$.019 \\
          						& 2     	& .400$\pm$.037 	& .377$\pm$.040 	& .357$\pm$.012 	& .287$\pm$.024 \\
          						& 3     	& .345$\pm$.048 	& .309$\pm$.058 	& .336$\pm$.012 	& .250$\pm$.018 \\
     \hline
    \multicolumn{1}{l|}{\multirow{3}[1]{*}{{\rc}}} 
    		  					& 1     	& .519$\pm$.028 	& .731$\pm$.027 	& .390$\pm$.014 	& .306$\pm$.023 \\
          						& 2     	& .469$\pm$.035 	& .666$\pm$.027 	& .371$\pm$.013 	& .288$\pm$.021 \\
          						& 3     	& .380$\pm$.048 	& .524$\pm$.034 	& .363$\pm$.010 	& .267$\pm$.020 \\
     \hline
    \multicolumn{1}{l|}{\multirow{3}[1]{*}{{\lws}}} 
    		  					& 1     	& .242$\pm$.042 	& .435$\pm$.049 	& .225$\pm$.038 	& .289$\pm$.017 \\
          						& 2     	& .239$\pm$.048 	& .406$\pm$.040	& .207$\pm$.034 	& .271$\pm$.014 \\
          						& 3     	& .218$\pm$.017 	& .318$\pm$.064	& .216$\pm$.029 	& .244$\pm$.023 \\
     \hline
    \multicolumn{1}{l|}{\multirow{3}[1]{*}{{\aggd}}} 
    		  					& 1     	& .527$\pm$.035 	& .391$\pm$.040 	& .383$\pm$.014 	& .397$\pm$.028 \\
          						& 2     	& .439$\pm$.020 	& .371$\pm$.037 	& .372$\pm$.020 	& .360$\pm$.029 \\
          						& 3     	& .321$\pm$.043 	& .327$\pm$.028 	& .344$\pm$.011 	& .328$\pm$.023 \\
     \hline  \hline
    \end{tabular}
       \caption{The classification accuracies (mean$\pm$std) of {\ours} and comparative algorithms on the benchmark datasets with varying numbers of false positive candidate labels ($r \in \{1,2,3\}$).}
        \label{tab:res_benckmark}
\end{table}
 
\subsection{Results on the Benchmark Datasets}
\label{subsec:res_benchmark}
Table \ref{tab:res_benckmark} presents the results of {\ours} and the comparative algorithms on benchmark datasets, considering varying numbers of false positive labels ($r \in \{1,2,3\}$). Compared to MIPL algorithms, {\ours} consistently achieves higher average accuracy than {\demipl} and {\miplgp}. Furthermore, in contrast to PLL algorithms, {\ours} significantly outperforms them in all cases.

For the MNIST-{\scriptsize{MIPL}} and FMNIST-{\scriptsize{MIPL}} datasets, each with $5$ class labels, {\ours} achieves an average accuracy at least $0.016$ higher than {\demipl} and between $0.032$ to $0.17$ higher than {\miplgp}. In the case of the Birdsong-{\scriptsize{MIPL}} dataset that comprises $13$ class labels, {\ours}'s average accuracy surpasses {\demipl} by at least $0.021$ and {\miplgp} by at least $0.055$. The SIVAL-{\scriptsize{MIPL}} dataset spans $25$ class labels, encompassing diverse categories such as fruits and commodities. {\ours}'s average accuracy surpasses {\demipl} by $0.04$ to $0.097$ and {\miplgp} by an average of $0.013$. Notably, {\demipl} demonstrates relatively superior performance with fewer class labels, while {\miplgp} excels in scenarios with more class labels. In contrast, {\ours} consistently maintains the highest average accuracy in both fewer and more class labels. This indicates that {\ours} exhibits superior capabilities compared to existing MIPL algorithms.
PLL algorithms exhibit decent results on the MNIST-{\scriptsize{MIPL}} and FMNIST-{\scriptsize{MIPL}} datasets when $r=1$ or $r=2$. However, their performance significantly deteriorates when $r=3$ or on the Birdsong-{\scriptsize{MIPL}} and SIVAL-{\scriptsize{MIPL}} datasets. This observation underscores the intrinsic complexity of MIPL problems, highlighting that they cannot be reduced to PLL problems. 

The above analysis not only highlights the robustness of {\ours} across diverse label space but also emphasizes the limitations of addressing MIPL problems using PLL algorithms. The results underscore the importance of algorithmic designs specifically tailored to MIPL tasks. 

\subsection{Results on the Real-World Datasets}
\label{subsec:res_real}
\begin{table}[!t]
  \centering  \scriptsize
    \begin{tabular}{p{2.cm} | cccc p{1.1cm}  p{1.1cm}  p{1.1cm}  p{1.1cm}}
    \hline  \hline
    \multicolumn{1}{l|}{Algorithm}  	& C-Row   			& C-SBN   			& C-KMeans  			& C-SIFT \\
    \hline
    \multicolumn{1}{l|}{{\ours}}  		& .433$\pm$.008 		& .509$\pm$.007 		& \textbf{.546$\pm$.012} 	& \textbf{.540$\pm$.010} \\
    \multicolumn{1}{l|}{{\demipl}}  	& .408$\pm$.010 		& .486$\pm$.014  		& .521$\pm$.012		& .532$\pm$.013 \\
    \multicolumn{1}{l|}{{\miplgp}}  	& .432$\pm$.005 		& .335$\pm$.006 		& .329$\pm$.012  	& -- \\
    \hline  
    \multicolumn{5}{c}{Mean} \\
    \hline
    \multicolumn{1}{l|}{{\proden}} 	& .365$\pm$.009 		& .392$\pm$.008 		& .233$\pm$.018 		& .334$\pm$.029 \\
    \multicolumn{1}{l|}{{\rc}}    		& .214$\pm$.011 		& .242$\pm$.012 		& .226$\pm$.009		& .209$\pm$.007 \\
    \multicolumn{1}{l|}{{\lws}}   		& .291$\pm$.010 		& .310$\pm$.006 		& .237$\pm$.008 		& .270$\pm$.007 \\
    \multicolumn{1}{l|}{{\aggd}} 		& .412$\pm$.008 		& .480$\pm$.005		& .358$\pm$.008		& .363$\pm$.012 \\
    \hline  
    \multicolumn{5}{c}{MaxMin} \\
    \hline  
    \multicolumn{1}{l|}{{\proden}} 	& .401$\pm$.007 	  	& .447$\pm$.011 		& .265$\pm$.027 		& .291$\pm$.011 \\
    \multicolumn{1}{l|}{{\rc}}    		& .227$\pm$.012 		& .338$\pm$.010 		& .208$\pm$.007		& .246$\pm$.008 \\
    \multicolumn{1}{l|}{{\lws}}   		& .299$\pm$.008 		& .382$\pm$.009 		& .247$\pm$.005 		& .230$\pm$.007 \\
    \multicolumn{1}{l|}{{\aggd}} 		& \textbf{.460$\pm$.008}	& \textbf{.524$\pm$.008} 	& .434$\pm$.009		& .285$\pm$.009 \\
    \hline  \hline
    \end{tabular}
      \caption{The classification accuracies (mean$\pm$std) of {\ours} and comparative algorithms on the real-world datasets.}       
\label{tab:res_crc}
\end{table}

\begin{figure*}[!t]
\centering
\begin{overpic}[width=140mm]{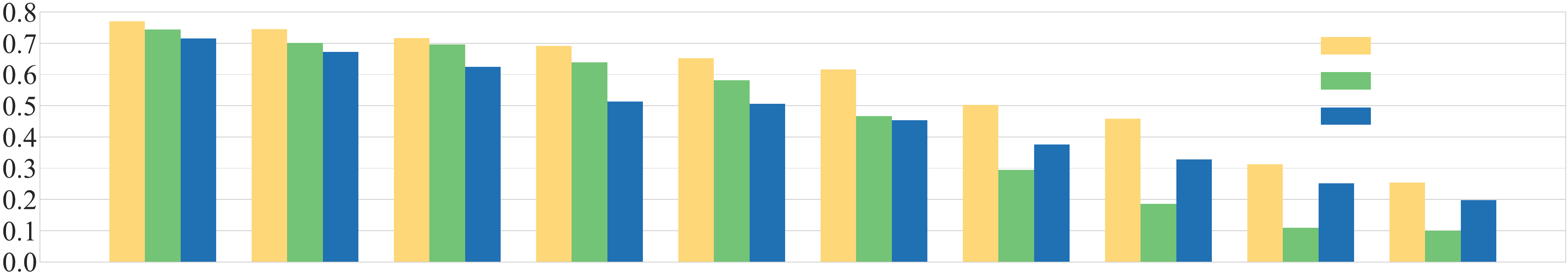}   
\put(-20, 52) {\small \rotatebox{90}{accuracy}} 
\put(880, 142) {\small {\ours}} 
\put(880, 119) {\small {\demipl}} 
\put(880, 97) {\small {\miplgp}} 
\put(78, -5) {\small  $r=1$} 
\put(167, -5) {\small  $r=2$} 
\put(259, -5) {\small  $r=3$} 
\put(350, -5) {\small  $r=4$} 
\put(442, -5) {\small  $r=5$} 
\put(532, -5) {\small  $r=6$} 
\put(623, -5) {\small  $r=7$} 
\put(714, -5) {\small  $r=8$} 
\put(800, -5) {\small  $r=9$} 
\put(888, -5) {\small  $r=10$} 
\end{overpic}
\caption{The classification accuracies of {\ours}, {\demipl}, and {\miplgp} on the Birdsong-{\scriptsize{MIPL}} dataset with varying $r$.} 
  \label{fig:birdsong_r}
\end{figure*}

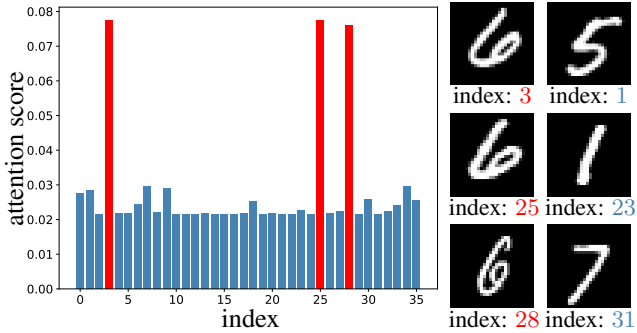
\begin{figure}[!t]
\begin{overpic}[width=57mm]{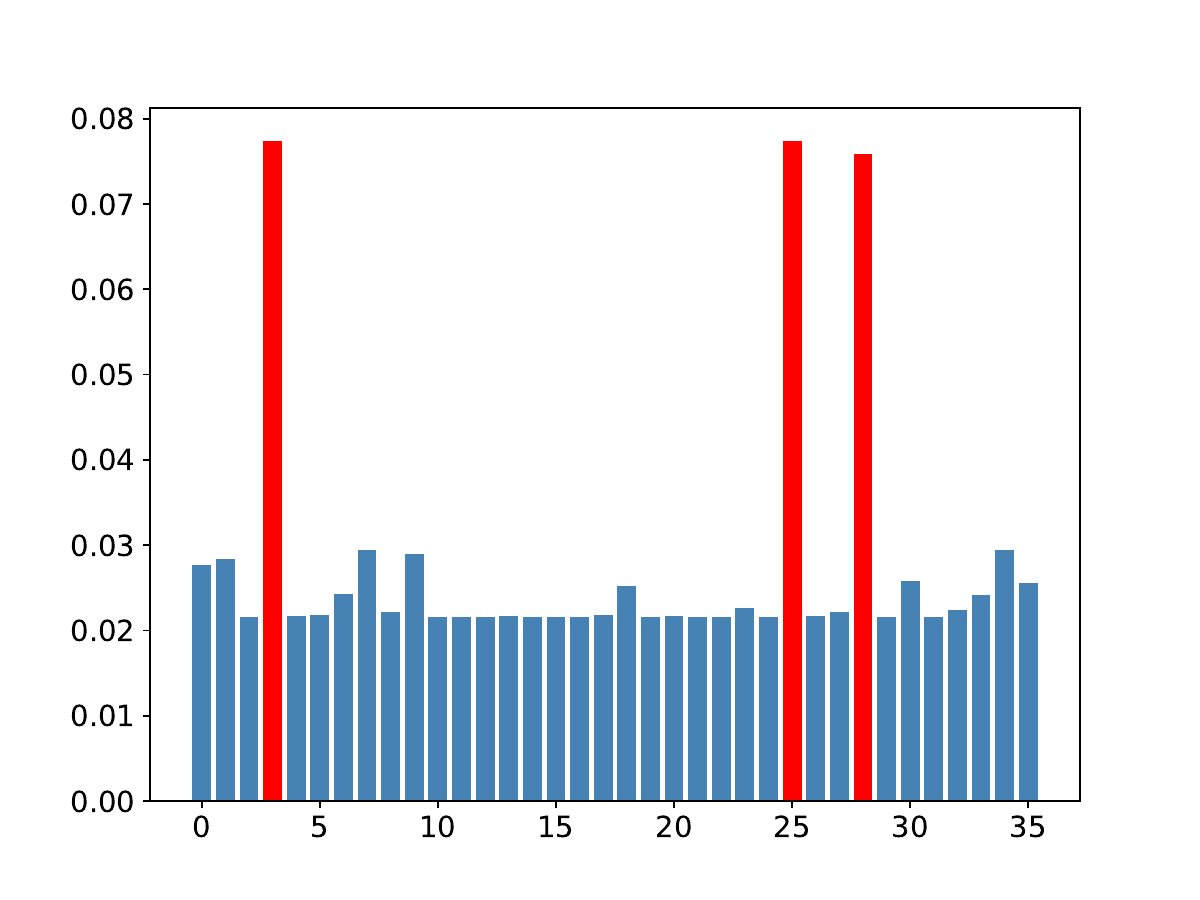}
  \put(-10, 220) {\rotatebox{90}{attention score}}
  \put(495, -15) {index} 
  \put(1030, 550){\includegraphics[width=11mm]{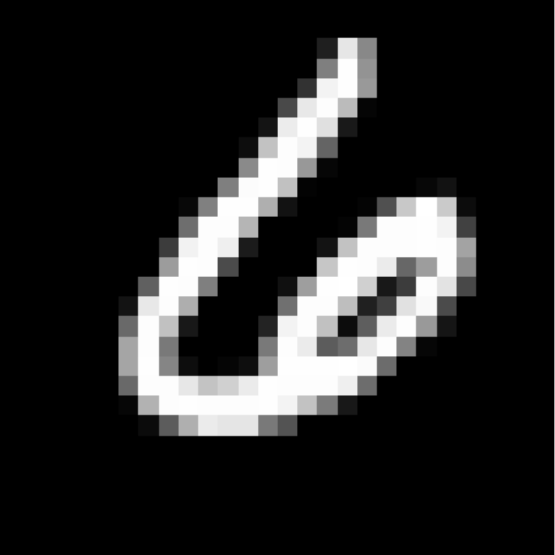}}
  \put(1035, 508) {\small index: \textcolor{red}{$3$}} 
  \put(1030, 290){\includegraphics[width=11mm]{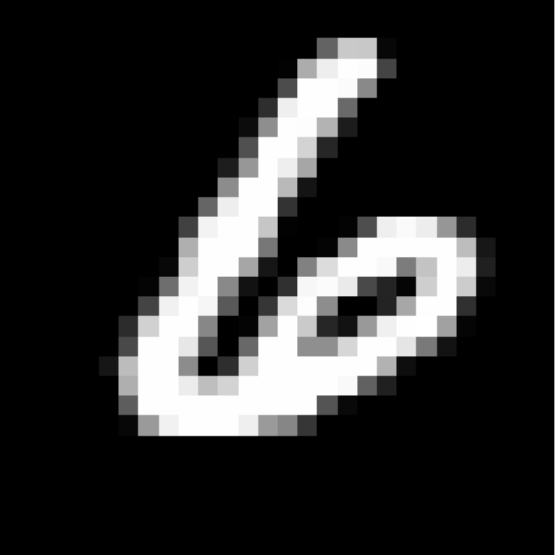}}
  \put(1025, 245) {\small index: \textcolor{red}{$25$}} 
  \put(1030, 30){\includegraphics[width=11mm]{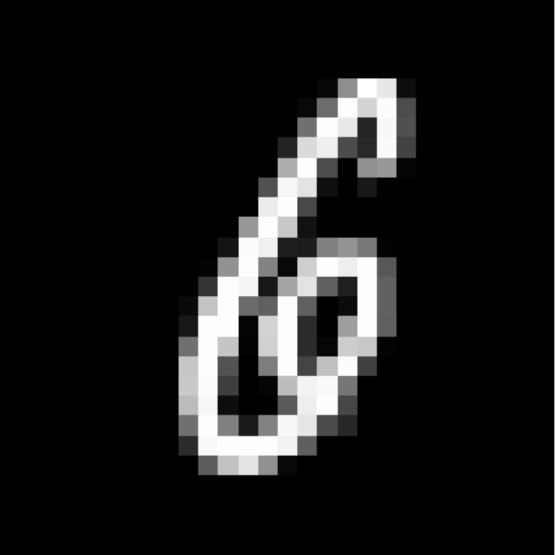}}
  \put(1025, -15) {\small index: \textcolor{red}{$28$}} 
  \put(1256, 550){\includegraphics[width=11mm]{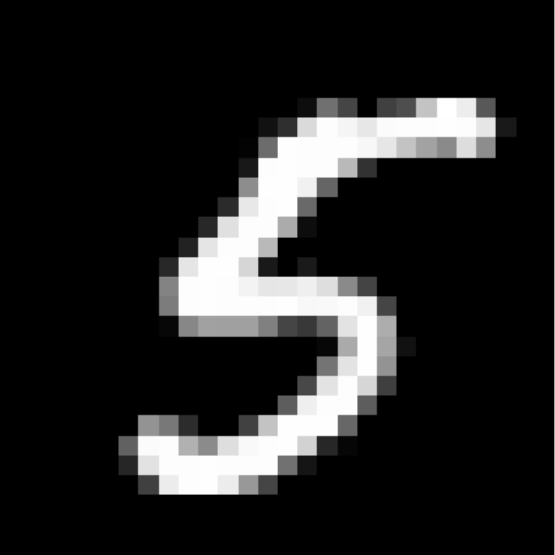}}
  \put(1260, 508) {\small index: \textcolor{steelblue}{$1$}} 
  \put(1255, 290){\includegraphics[width=11mm]{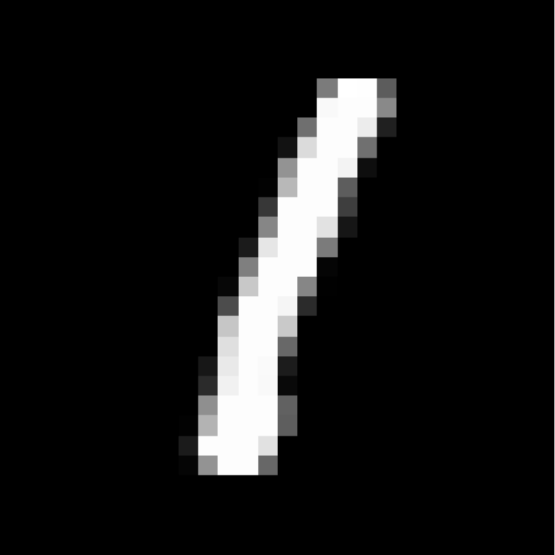}}
  \put(1250, 245) {\small index: \textcolor{steelblue}{$23$}} 
  \put(1255, 30){\includegraphics[width=11mm]{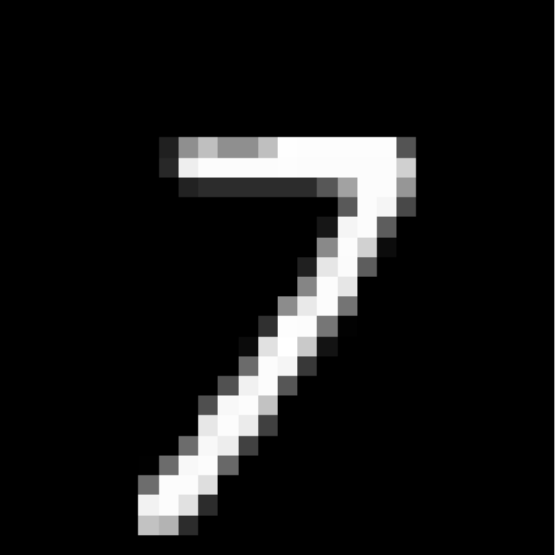}}
  \put(1250, -15) {\small index: \textcolor{steelblue}{$31$}} 
\end{overpic}
 \caption{Attention scores for a test bag. Red and blue are the attention scores of positive and negative instances, respectively. }
\label{fig:visual}
\end{figure}

Table \ref{tab:res_crc} provides the results of {\ours} and the comparative algorithms on the CRC-{\scriptsize{MIPL}} dataset. The symbol -- denotes cases where results could not be obtained due to memory overflow on our server. Compared to MIPL algorithms, {\ours} consistently achieves the highest average accuracies. Additionally, in comparison to PLL algorithms, {\ours} significantly outperforms them in $30$ out of $32$ cases.

For the CRC-{\scriptsize{MIPL}} dataset, both {\ours} and {\demipl} exhibit improved performance as the complexity of the image bag generator increases. This observation aligns with the intuition that, while avoiding overfitting, intricate feature extractors tend to produce higher classification accuracies. However, this phenomenon is not consistently observed for {\miplgp} and PLL algorithms. For example, these algorithms do not consistently achieve superior results on the C-KMeans and C-SIFT datasets compared to the results on the CRC-Row or C-SBN dataset. We posit that the intricate features exceed the capability limits of these algorithms. Thus, the development of effective MIPL algorithms becomes imperative.

In most cases, the MaxMin strategy tends to yield superior outcomes than the Mean strategy. We postulate that this difference arises from the significant distinction between tissue cells and the background in the CRC-{\scriptsize{MIPL}} dataset. Applying the Mean strategy to features generated by simple image bag generators (i.e., Row and SBN) diminishes the distinction between tissue cells and the background, making it challenging to learn discriminative features. Conversely, for features generated by more complex image bag generators (i.e., KMeansSeg and SIFT), both the Mean and MaxMin strategies demonstrate their respective merits. Therefore, both strategies are worthy of consideration and application.

\begin{table}[!t]
  \centering  
  \scriptsize 
   \begin{tabular}{l|c|cccc}
    \hline  \hline
    \multicolumn{1}{l|}{Dataset} 		& $r$ 		& {\ours} 				& {\oursma}+{\ourssp} 	& {\oursma}+{\oursin}	& {\oursma} \\
    \hline
    \multicolumn{1}{l|}{\multirow{3}[1]{*}{Birdsong}}
    							& 1     		& .771$\pm$.018 		& .742$\pm$.014 		& .746$\pm$.015 		& .733$\pm$.011 \\
          						& 2     		& .745$\pm$.015 		& .665$\pm$.024 		& .689$\pm$.020 		& .677$\pm$.017 \\
          						& 3     		& .717$\pm$.017 		& .592$\pm$.031 		& .674$\pm$.023 		& .652$\pm$.016 \\
	\hline
	\multicolumn{1}{l|}{\multirow{3}[1]{*}{SIVAL}}
    							& 1     		& .675$\pm$.022 		& .618$\pm$.021 		& .626$\pm$.019 		& .620$\pm$.022 \\
          						& 2     		& .616$\pm$.025 		& .532$\pm$.041 		& .550$\pm$.040 		& .540$\pm$.038 \\
          						& 3     		& .600$\pm$.029 		& .545$\pm$.027 		& .521$\pm$.025 		& .521$\pm$.032 \\
     \hline  \hline
    \end{tabular}
     \caption{The classification accuracies of the variants on the Birdsong-{\scriptsize{MIPL}} and SIVAL-{\scriptsize{MIPL}} datasets. }
     \label{tab:crc}
\end{table}

\subsection{Further Analyses}
\label{subsec:further}
\textbf{Effectiveness of CLI.}
To evaluate the impact of CLI, we modify the loss function in Equation (\ref{eq:full_loss}) and propose three variants: {\oursma}+{\ourssp}, {\oursma}+{\oursin}, and {\oursma}. These variants respectively represent the removal of inhibition loss, sparsity loss, and the simultaneous elimination of both inhibition and sparsity losses. Table \ref{tab:crc} presents the experimental results conducted on the Birdsong-{\scriptsize{MIPL}} and SIVAL-{\scriptsize{MIPL}} datasets. With {\oursma} as the baseline, the introduction of individual sparse loss or inhibition loss tends to yield marginal performance improvements in most cases, while in some cases, performance degradation may occur. In contrast, {\ours}, using the CLI demonstrates a substantial boost in classification accuracy. 

\textbf{Challenging Disambiguation Scenarios.}
We select different quantities of false positive labels from $1$ to $10$ on the Birdsong-{\scriptsize{MIPL}} dataset. Figure \ref{fig:birdsong_r} presents the experimental results of {\ours}, {\demipl}, and {\miplgp} with $r \in \{1,2, \cdots,10\}$. Particularly, {\ours} and {\demipl} adhere to the embedded-space paradigm, while {\miplgp} follows the instance-space paradigm. Three distinct phenomena are observed: (a) {\ours} consistently exhibits higher average accuracy compared to {\demipl} and {\miplgp}. (b) For $r < 7$, {\demipl} outperforms {\miplgp}. However, when $r \ge 7$, {\miplgp} surpasses {\demipl}. (c) The gaps between {\ours} and {\demipl} are greater when $r \in \{6,7,8,9,10\}$ than when $r \in \{1,2,3,4,5\}$. The widening gaps signify the growing significance of the scaled additive attention mechanism and CLI. Therefore, Figure \ref{fig:birdsong_r} clearly demonstrates that {\ours} outperforms both {\miplgp} and {\demipl} in disambiguation, even when confronted with challenging scenarios.

\begin{table}[!t]
  \centering  
  \scriptsize 
     \begin{tabular}{p{1.3cm} |c|cccc p{0.8cm}  p{136cm}  p{1.1cm}  p{1.1cm} p{1.1cm} }
    \hline  \hline
    \multicolumn{1}{l|}{Algorithm} 	& $r$ 	& MNIST 				& FMNIST 			& Birdsong 			& SIVAL \\
    \hline
    \multicolumn{1}{l|}{\multirow{3}[1]{*}{CLI loss}}
    							& 1     	& .992$\pm$.007 		& .903$\pm$.018 		& .771$\pm$.018		& .675$\pm$.022 \\
          						& 2  		& .987$\pm$.010 		& .845$\pm$.026 		& .745$\pm$.015 		& .616$\pm$.025 \\
          						& 3     	& .748$\pm$.144 		& .702$\pm$.055 		& .717$\pm$.017 		& .600$\pm$.029 \\
     \hline
    \multicolumn{1}{l|}{\multirow{3}[1]{*}{ \textsc{Ce-Sp-In} }}
    							& 1     	& .899$\pm$.037 		& .825$\pm$.035 		& .740$\pm$.013 		& .639$\pm$.030 \\
          						& 2    	& .847$\pm$.027 		& .679$\pm$.037		& .687$\pm$.024 		& .587$\pm$.022 \\
          						& 3     	& .636$\pm$.112		& .610$\pm$.037 		& .592$\pm$.036	 	& .578$\pm$.022 \\
     \hline
   	\multicolumn{1}{l|}{\multirow{3}[1]{*}{\textsc{Ce}}} 
							& 1		& .919$\pm$.017 		& .709$\pm$.257 		& .704$\pm$.019 		& .587$\pm$.028 \\
							& 2		& .833$\pm$.016 		& .645$\pm$.044 		& .616$\pm$.032 		& .534$\pm$.025 \\
							& 3		& .628$\pm$.096 		& .551$\pm$.032		& .459$\pm$.045 		& .514$\pm$.025\\
     \hline  \hline
    \end{tabular}
       \caption{\small The comparison between the CLI loss and the CE loss. }
        \label{tab:res_ce}
\end{table}

\textbf{Comparison of CLI and Cross-Entropy Loss.}
For a comparative analysis between the CLI loss and the cross-entropy (CE) loss, we substitute the mapping loss and the CLI loss with the CE loss, resulting in variants \textsc{Ce-Sp-In} (which utilizes CE loss, sparsity loss, and inhibition loss) and \textsc{Ce} (which only utilizes CE loss), respectively. 
Table \ref{tab:res_ce} illustrates that, in all cases, accuracies obtained with the CLI loss surpass those achieved with \textsc{Ce-Sp-In} and \textsc{Ce}. Notably, the incorporation of inhibition loss and sparsity loss enhances the performance of the CE loss, underscoring the importance of considering the intrinsic properties of the label space and the information from non-candidate label sets.

\textbf{Interpretability of the Attention Mechanism.}
Figure \ref{fig:visual} displays the attention scores of a test multi-instance bag in the MNIST-{\scriptsize{MIPL}} dataset ($r=1$). The bag contains positive instances represented by the digit $6$, while negative instances are drawn from digits $\{1, 3, 5, 7, 9\}$. Additionally, we visualize the attention scores of all three positive instances and the negative instances. Figure \ref{fig:visual}  illustrates that {\ours} can accurately identify all positive instances by assigning significantly higher attention scores to them, and the attention scores can be directly utilized for interpretability. 

\section{Conclusion}
\label{sec:con}
This paper investigates a multi-instance partial-label learning algorithm that introduces a scaled additive attention mechanism and exploits conjugate label information. This information includes both candidate label information and non-candidate label information, along with the sparsity of the true label matrix. Experimental results demonstrate the superiority of our proposed {\ours} algorithm. The utilization of CLI proves significantly more effective than relying on incomplete label information, especially in challenging disambiguation scenarios. 
In the future, we will explore the instance-depend MIPL algorithm and conduct theoretical analyses to develop more effective algorithms.

\bibliographystyle{named}
\bibliography{myref}

\end{document}